\documentclass[]{ijimai_article}

\usepackage{amsmath} 
\usepackage{booktabs}
\usepackage{lipsum}
\usepackage{float}
\usepackage{hyperref}

\title{Virtual Trial Room with Computer Vision and Machine Learning}
\author{Tulashi Prasad Joshi \and Amrendra Kumar Yadav \and Arjun Chhetri \and Suraj Agrahari \and Umesh Kanta Ghimire}
\affiliation{Tribhuwan University,
Institute of Engineering, Thapathali Campus, Kathmandu (Nepal)}

\mailinfo{E-mail address: tha075bei047@tcioe.edu.np (Tulashi Prasad Joshi), tha075bei005@tcioe.edu.np (Amrendra Kumar Yadav), tha075bei007@tcioe.edu.np (Arjun Chhetri), tha075bei043@tcioe.edu.np (Suraj Agrahari), ukg@tcioe.edu.np (Umesh Kanta Ghimire).}

\abstract{Online shopping has revolutionized the retail industry, providing customers with convenience and accessibility. However, customers often hesitate to purchase wearable products such as watches, jewelry, glasses, shoes, and clothes due to the lack of certainty regarding fit and suitability. This leads to significant return rates, causing problems for both customers and vendors. To address this issue, a platform called the Virtual Trial Room with Computer Vision and Machine Learning is designed which enables customers to easily check whether a product will fit and suit them or not. To achieve this, an AI-generated 3D model of the human head was created from a single 2D image using the DECA\cite{ref1} model. This 3D model was then superimposed with a custom-made 3D model of glass which is based on real-world measurements and fitted over the human head. To replicate the real-world look and feel, the model was retouched with textures, lightness, and smoothness. Furthermore, a full-stack application was developed utilizing various fornt-end and back-end technologies. This application enables users to view 3D-generated results on the website, providing an immersive and interactive experience.}

\begin{document}

\maketitle
\section{INTRODUCTION}
\fancyFirstWord{The} rapid growth of technology, particularly through smart devices and online platforms, has revolutionized the way people shop, enabling consumers to browse and purchase products from anywhere at any time. While this shift has made transactions more convenient, hesitation remains when buying wearable products such as jewelry, clothing, shoes, and accessories, as customers cannot physically try them on. This leads to higher return rates, which pose significant challenges for retailers, resulting in lost time, increased operational costs, and potential fraudulent returns. To address these issues, this research focuses on developing a 3D virtual trial machine that allows users to virtually try on wearable products using a personalized digital avatar. By employing the DECA (Deeply Embedded 3D Face Reconstruction) model, the study creates accurate 3D reconstructions of users' faces from 2D images, enabling them to see how products like jewelry or eyewear will look on their own face. Additionally, the system fits virtual products, such as glasses or accessories, seamlessly onto the reconstructed 3D model, providing a realistic preview of how these items will look in real life. The primary objectives of this research are to enhance customer confidence in online purchases, reduce product return rates, and minimize the occurrence of fraudulent returns, ultimately providing a more efficient, secure, and satisfying shopping experience for both consumers and sellers.

\section{RELATED WORK}\label{struct}
There have been numerous endeavors for virtual trial applications. Before 3D trial rooms and 3D reconstruction,
Shreya et al, \cite{ref2} "Virtual trial room using augmented reality", explained the concept of Virtual trial using Augmented Reality. Augmented Reality is direct and indirect view of real-world elements that are augmented on computer. Augmented Reality considers real and Virtual elements. This method used a library called OpenCV for image detection and processing. The image was captured and converted into 3D model and then the person can change the logo and color of the cloths that s/he is buying. But for the early phase this project was implemented on static photos but later with the advancement of project a video capturing device was used and that video was used to form the 3D image of person. \\

Isikdogan et al., \cite{ref3} “A Real Time Virtual Dressing Room Application using Kinect”, introduced a virtual trial room using Microsoft Kinect sensor. This method extracted the user from a video stream provided to it, the Kinect sensor then detected the depth of the object. Similarly, the cloth was also scanned and then the user skin and cloths were layered together to get the output. This reported an average percentage of overlap between the user and the cloth models of 83.97\%. Though this project was for 2D but the motion detection technique allowed to check the cloths in real time without the need to wear the cloths in the spot.\\

Nikki et al., \cite{ref4}” Virtual Trial Room” Vishwakarma Journal of Engineering Research, Volume 1 Issue 4, December 2021 proposed a human friendly interface to check the clothes Virtually without needing to manually try each of them. This method divided the whole project into 3 phases: first was detection and sizing of the customers body structure, second was detection of reference points based on the location and size of face and upper body points i.e., the shoulder points and finally the third was superimposition of the clothing over the customers' image. In contrast to other VTR techniques, the major difference was being economical instead of using heavy and huge costing hardware components or peripherals. The proposed technique was software based and was designed to be universally compatible as long as the system has a camera.\\

Vipin et al., \cite{ref5} “VIRTUAL TRIAL ROOM”, They decided to build an Online Trial Room Application to eliminate the problem faced by both the customers and vendors regarding the size of the clothes. Their research was based on the concept of taking a video from the user using a camera device and then the video is divided into individual frames. The frames are then used to extract the user’s body. Finally using the junctions to exact the information on the placement of joints in the body to perform the transformation, rotation and scale the wearable image onto the user in real-time. This project was implemented in Flask Web application with OpenCV a python\cite{ref6} Module. This application worked on devices with an inbuilt or attached camera, internet, and web browser. \\

Davide et al., \cite{ref7} “Designing an AI-Based Virtual Try-On Web Application" is an eyewear Virtual try-on based on a framework that leverages advanced deep learning based on computer vision techniques. They started by studying and understanding the underlying architecture, components, and their interactions in designing an 3D model of the head. They studied different predesigned models and used extensive analysis and experiments for evaluating their design, complexity, geometry reconstruction errors and reconstructed texture quality. This allowed them to select best suited model for them to use. After this evaluation they used PRNet\cite{ref8} as the method for 3D face reconstruction for their Virtual try-on. Their product actual glassed and face size to provide realistic fit. Finally, the user could interact with the system by using a web application optimized for desktop and mobile devices. At the end they perform a usability study that showed above-average score. But this method had some limitation such as the quality of the 3D model was not so good and the texture quality was compromised. Also, the 3D model had some missing portions(holes) in it.\\

Arthur et al.,\cite{ref9} “Virtual Try-On of Eyeglasses Using 3d Model of The Head”, this work presented a system for Virtual try-on of eyeglasses using a 3D model of user’s face and head. The 3D head model was reconstructed using only one image of the user. The 3D glasses model is then fitted onto this head model, and the user’s head movement was tracked in real-time to rotate the 3D head model with glasses accordingly. The main advantages of this system were that the head can be moved i.e., rotated to certain angle so the user could see how it looked from the other sides. Movement parameters were used to rotate the head model at about 90 degrees. A non-linearity was introduced in the real head movement to the movement of the head model.  \\

Amit et al., \cite{ref10} “Online Virtual Trial Room Implementation Using OpenCV Python” designed a system based on Open-Source Computer Vision Library to allow the customers view whether the wearable would suit them or not. This project took the video and divided it by frames by frame. The project uses OpenCV, this is a light weight library of python. The OpenCV allowed the detection of body parts from the video frames. Once the system recognized the body parts then it superimposed the clothes on the top of the detected body parts. The major success of this project was based on two factors first is accuracy of superposition and second of the realistic view of the Virtual Trial Room. The accuracy of superposition is the ability of the program to detect the user body parts in the frames. There are two ways of doing that first is the use of neural network to determine the human body parts in the frame, and other is the use of marker like a BGR. This uses the pixels to determine the user from the frame. The realistic view of the user is not possible but the program tries to show the user as the user would see him/herself in the mirror. \\ 

Anagaha et al., \cite{ref11} “3D Virtual Trial Room” AR (Augmented Reality) showed the direct or indirect view of the real-world element. This project used the AR technology to allow the users to try on different colours of clothes. The user could change the colour of the wearable and change the logo. Aswini et al., \cite{ref12} “A REAL TIME VIRTUAL DRESSING ROOM APPLICATION USING OPENCV” proposed a system that could take user image and then superimpose a cloth on them.

\section{DATASET}
DECA is trained on three publicly available datasets: VGGFace2\cite{ref13}, BUPT-Balancedface\cite{ref14} and Vox-Celeb2\cite{ref15, ref16, ref17}VGGFace2 contains images of over 8000 subjects, with an average of more than 350 images per subject. BUPT-Balancedface offers 7000 subjects per ethnicity (i.e., Caucasian, Indian, Asian and African), and VoxCeleb2 contains 145000 videos of 6000 subjects. In total, DECA is trained on 2 million images. All datasets are provided with an identity label with different facial expressions and annotated with the facial landmarks and emotion.
To evaluate the performance of the DECA model, two benchmark datasets were used: the NoW challenge[15] dataset and the Feng et al. low-quality and high-quality face image datasets. These datasets contain 3D face models and corresponding 2D images captured under different lighting, pose, and expression conditions. The DECA model was shown to achieve state-of-the-art performance on both benchmarks.
The NoW (Nose on Wall) challenge dataset is a publicly available dataset for evaluating face reconstruction algorithms under unconstrained imaging conditions. The dataset contains 1,000 high-resolution (2,000×2,000) RGB images of 53 subjects captured under varying lighting conditions, poses, and expressions. Each image is annotated with 68 facial landmarks. This dataset was used to evaluate the performance of DECA in reconstructing faces under unconstrained imaging conditions.
The Feng et al. low-quality and high-quality face image datasets were created to evaluate the performance of face reconstruction algorithms under different image qualities. The low-quality dataset contains 300 images with low resolution, motion blur, and out-of-focus blur, while the high-quality dataset contains 300 images with high resolution and clear focus. These datasets were used to evaluate the robustness of DECA to low-quality input images.

\section{IMPLEMENTATION}

\begin{figure}[h]
    \centering
    \includegraphics[width=1\linewidth]{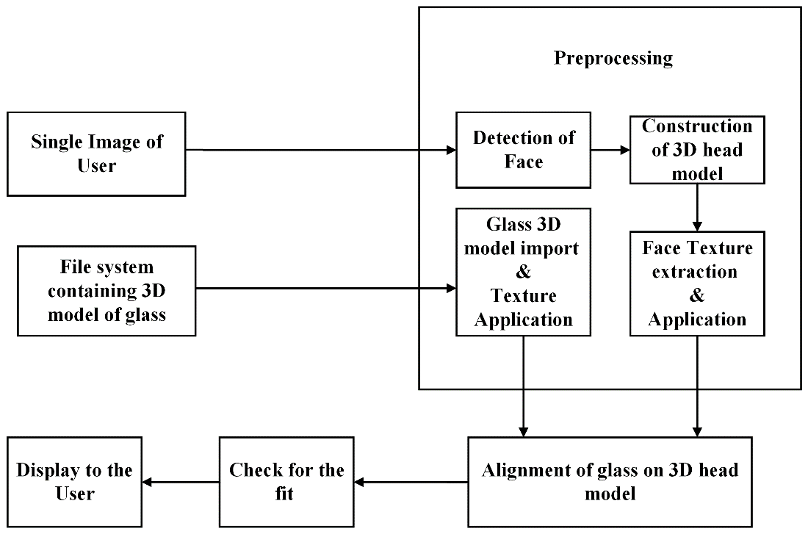}
    \caption{System Block Diagram}
    \label{fig:enter-label}
\end{figure}
This system implements a platform to try out eyeglasses eliminating the need to go to stores in person. Initially, a high resolution image of a person is taken. This image goes through pre-processing which involves steps like face detection, face reconstruction, texture extraction etc. The custom 3D models of the glasses are stored somewhere in the file system. These 3D eyeglasses are placed on the 3D model of human head and aligned properly. The texture is applied to both glasses and head model and superimposed model is exported to the file system. This superimposed model is then presented to the customers through website where they can check the appearance and fit in a photorealistic 3D reconstructed model.
\subsubsection{DECA Model}
DECA stands for Detailed Expression Capture and Animation, and it refers to a technology or technique used to create highly detailed and accurate digital representations of human faces and bodies, including facial expressions and body movements. The process of DECA typically involves capturing detailed data or images of the human face and body, either through 3D scanning or by using cameras. The captured data is then processed to create highly accurate digital models of the face and body, which can be animated in real-time to simulate complex facial expressions, body movements, and other gestures.\\
Monocular 3D face reconstruction methods can recover fine geometric details, but it has several limitations like inability to realistically animate faces due to a lack of modeling of how wrinkles vary with expression, poor generalization to in-the-wild images as they are often trained on high-quality face scans and limited ability to capture and separate person-specific details from expression-dependent wrinkles. DECA model addresses these limitations of current monocular 3D face reconstruction methods. DECA consists of two parts: a UV displacement map generator that regresses person-specific detail and expression parameters from a low-dimensional latent representation and a regressor that predicts detail, shape, albedo, expression, pose, and illumination from a single image. To achieve this, a novel detail-consistency loss that separates person-specific details from expression-dependent wrinkles is introduced. This separation allows us to synthesize realistic wrinkles while keeping person-specific details unchanged. DECA is trained on in-the-wild images with no 3D supervision and outperforms current methods in terms of shape reconstruction accuracy. Results on real-world data show the robustness and animation capabilities of DECA due to its ability to disentangle identity- and expression-dependent details. 

\begin{figure}[h]
    \centering
    \includegraphics[width=1\linewidth]{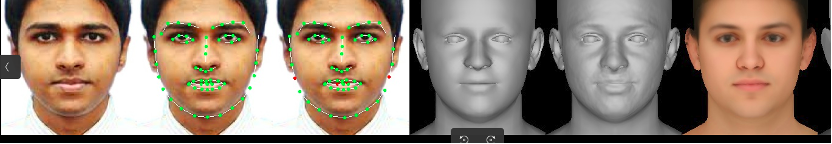}
    \caption{Series of markers placed on an actor’s face to track their facial movements}
    \label{fig:enter-label}
\end{figure}

\begin{figure}[h]
    \centering
    \includegraphics[width=1\linewidth]{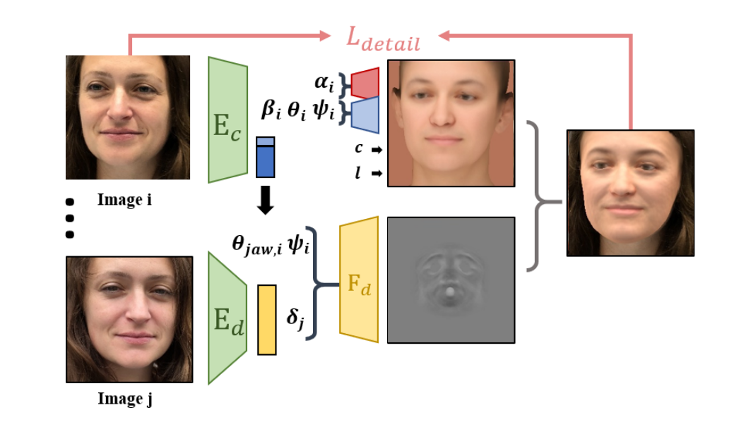}
    \caption{Detail consistency loss}
    \label{fig:enter-label}
\end{figure}
In this project DECA model is mainly used to create 3D head from single 2D image. 3D head is made up of face and coarse. DECA reconstructs the 3D face shape with mid-frequency geometric details and coarse with the help of FLAME’s model\cite{ref18}.

The detail consistency loss is defined as:

\begin{equation}
L_{\text{dc}} = L_{\text{detail}} \left( I_i, R \left( M (\beta_i, \theta_i, \psi_i), A(\alpha_i), F_d (\delta_j, \psi_i, \theta_{\text{jaw},i}), l_i, c_i \right) \right)
\end{equation}

where:
\begin{itemize}
    \item \( I_i \) and \( I_j \) are two images of the same subject.
    \item \( M (\beta_i, \theta_i, \psi_i) \) is a function that combines the parameters \( \beta_i \), \( \theta_i \), and \( \psi_i \) to produce the rendering.
    \item \( A(\alpha_i) \) represents the appearance of the face.
    \item \( F_d (\delta_j, \psi_i, \theta_{\text{jaw},i}) \) is a function that combines the detail code \( \delta_j \) and the parameters \( \psi_i \) and \( \theta_{\text{jaw},i} \) to produce the wrinkle detail.
    \item \( l_i \) and \( c_i \) are additional parameters.
    \item \( R \) is a function that produces the final rendering.
\end{itemize}
\subsubsection{FLAME Model}
The FLAME model is a lightweight and expressive generic head model that has been learned from over 33,000 accurately aligned 3D scans. The trained 3D face models and registrations for the dynamic D3DFACS dataset are provided, as well as demo code in Chumpy\cite{ref19} and Tensorflow\cite{ref20}. The code can be used to load and sample the model, and to fit the model to 3D landmarks.
The field of 3D face modeling presents a challenge in balancing between highly realistic, yet manual-labor intensive, methods and low-end methods that lack expressiveness. This FLAME (Faces Learned with an Articulated Model and Expressions) model aims to bridge this gap by using machine learning to generate a high-quality 3D face model from thousands of accurately aligned 3D scans. This model is designed to be compatible with existing graphics software and easy to fit to new data. FLAME uses a linear shape space that is trained from 3800 scans of human heads and features an articulated jaw, neck, and eyes, pose-dependent corrective blend shapes, and global expression blend shapes. These articulations are learned from 4D face sequences in the D3DFACS dataset and additional 4D sequences. The template mesh of the model is registered to the scan sequences and made available for research purposes. The FLAME model is low-dimensional but more expressive than existing models, such as FaceWarehouse\cite{ref21} and the Basel Face Model\cite{ref22}, as shown through comparison studies of fitting to both static 3D scans and 4D sequences using the same optimization method. With its high accuracy and accessibility for research purposes, FLAME offers a new approach to generating realistic 3D face models.
\subsubsection{Encoding and Decoding}
\textbf{Encoding:} The encoding process is performed by an encoder network \( E_c \), which can be mathematically represented as:
\begin{equation}
z = E_c(I)
\end{equation}
where \( z \) is the latent code, consisting of:
\begin{itemize}
    \item FLAME parameters for coarse geometry (\( \beta \), \( \psi \), \( \theta \)),
    \item albedo coefficients (\( \alpha \)),
    \item camera parameters (\( c \)), and
    \item lighting parameters (\( l \)).
\end{itemize}
In total, \( z \) has a dimensionality of 236.

\textbf{Decoding:} The decoding process is performed by a decoder network \( D \), which synthesizes the 2D image \( I_r \) based on the latent code:
\begin{equation}
I_r = D(z)
\end{equation}

The goal is to minimize the difference between the synthesized image and the input image, represented by the loss function \( L \):
\begin{equation}
L = \|I - I_r\|^2
\end{equation}

The encoding and decoding networks are trained together to minimize the loss function \( L \). This process results in a learned coarse reconstruction of the 3D face model from the input 2D image. The coarse reconstruction branch in the DECA method is trained by minimizing a loss function, \( L_{\text{coarse}} \), which is the sum of six different terms: 
\begin{itemize}
    \item Landmark loss \( L_{\text{lmk}} \),
    \item Eye closure loss \( L_{\text{eye}} \),
    \item Photometric loss \( L_{\text{pho}} \),
    \item Identity loss \( L_{\text{id}} \),
    \item Shape consistency loss \( L_{\text{sc}} \), and
    \item Regularization \( L_{\text{reg}} \).
\end{itemize}
\subsubsection{Eyeglss Modeling}
In this project,3D model of glass is created using 3D modeling software i.e., blender\cite{ref23}. There are several other software options available for creating 3D models, such as Maya, 3D Studio Max, SketchUp, etc.  
This process involves several steps, including modeling the basic shape of the glass, creating the glass material, and textures to give the glass its realistic appearance.
In Blender, the modeling process starts by creating a basic shape of the glass using geometry modeling tools, such as the Bezier curve, spline curve, or a simple cube. The shape can then be manipulated using various tools, such as scaling, rotating, or extruding to form the desired glass shape. Once the basic shape of the glass is created, the glass material can be added to the model. This involves creating a glass shader that simulates the way light travels through a glass object. The shader includes parameters for controlling the color, transparency, and reflections of the glass.
Finally, the glass model can be rendered to see the final result. To achieve a more realistic look, various techniques, such as adding reflections and refractions, can be used to enhance the appearance of the glass. This can be done by adjusting the reflectivity and refractive index properties of the glass shader. In this way Blender was used to construct the 3D glass.

\subsubsection{Glass Fitting}
The major concern of this project was the fitting of the glasses on the 3D head model of the user. There were different kinds of approaches such as manual fitting, fitting with scripts and using libraries. The idea was of putting the glasses at some distance from the eyes was used. However, we encountered another problem of separating out the eyes from the 3D Head model. We used blender python library that enabled the separation of eyeballs from the 3D head model. Once the eyeball was separated, the glasses were placed at a certain distance from the center of the eyes. It can be considered that the placement of the glasses from the center of the eyeball is almost the same for people of the same age group since the size of human eyes is almost the same for all humans of any age and gender.
The UV unwrapping of the 3D model is done in the blender. The texture is mapped with the unwrapped model and finally the 3D model with texture applied is generated. In the same way the texture is applied to the glass. The fitted 3D models are exported as glb files to for user visualization.
\begin{figure}[h]
    \centering
    \includegraphics[width=0.4\linewidth]{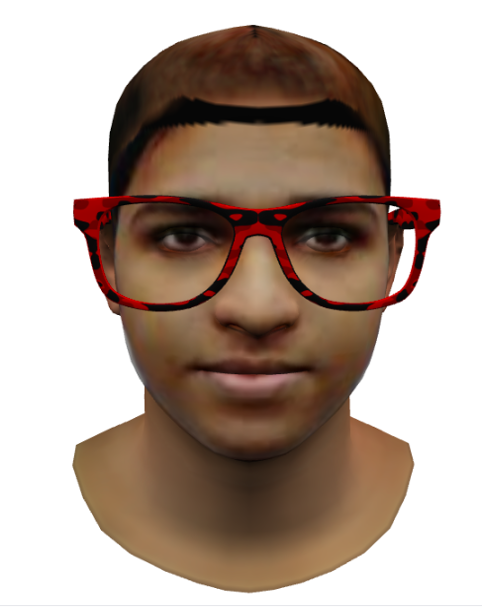}
    \caption{Glasses fitted on reconstructed face}
    \label{fig:enter-label}
\end{figure}
\section{RESULTS}\label{struct}
DECA is compared with other publicly available methods for 3D face reconstruction, including 3DDFA V2\cite{ref24, ref25}, Deng et al., RingNet\cite{ref26}, PRNet, 3DMM-CNN\cite{ref27}, and Extreme3D\cite{ref28}. The evaluation of the DECA method is compared to publicly available methods on two benchmark datasets: Now challenge[15] and Feng et al. challenge. The evaluation focuses on the accuracy of the coarse shape because there is no benchmark face dataset with ground truth shape details.
\begin{figure}[h]
    \centering
    \includegraphics[width=1\linewidth]{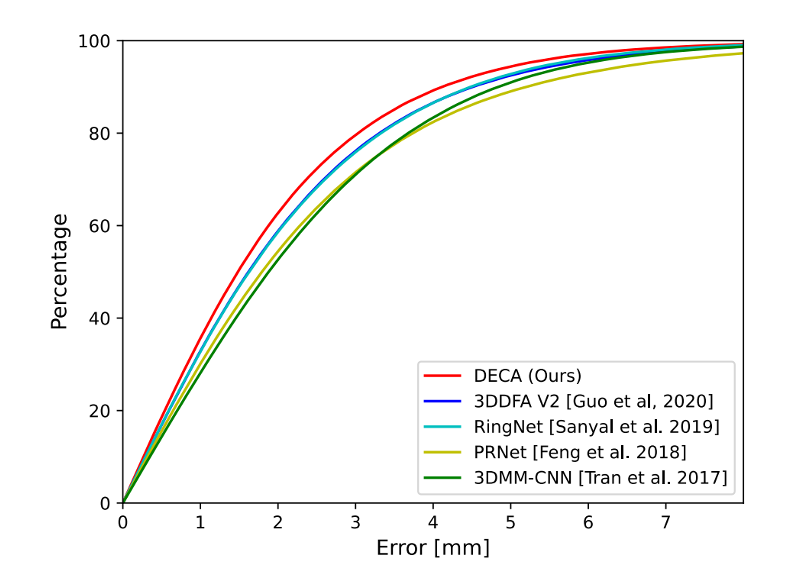}
    \caption{NoW challenge analysis curve}
    \label{fig:enter-label}
\end{figure}
The results show that DECA provides the best 3D face reconstruction performance compared to other methods, with the lowest mean, median, and standard deviation
\begin{table}[h]
    \centering
    \caption{Feng et al. benchmark performance. (Yao Feng, Feb 22)}
    \scriptsize 
    \begin{tabular}{@{}lcccccc@{}}
        \toprule
        \textbf{Method} & \textbf{Median (mm)} & \textbf{Mean (mm)} & \textbf{Std (mm)} \\
        & \textbf{LQ} & \textbf{HQ} & \textbf{LQ} & \textbf{HQ} & \textbf{LQ} & \textbf{HQ} \\ 
        \midrule
        3DMM-CNN     & 1.88 & 1.85 & 2.32 & 2.29 & 1.89 & 1.88 \\
        Extreme3D    & 2.40 & 2.37 & 3.94 & 3.58 & 6.15 & 6.75 \\
        PRNet        & 1.79 & 1.59 & 2.38 & 2.06 & 2.19 & 1.79 \\
        RingNet      & 1.63 & 1.59 & 2.08 & 2.02 & 1.79 & 1.69 \\
        3DDFA-V2     & 1.62 & 1.49 & 2.10 & 1.91 & 1.87 & 1.64 \\
        DECA         & 1.48 & 1.45 & 1.91 & 1.89 & 1.66 & 1.68 \\
        \bottomrule
    \end{tabular}
    \label{tab:3d_face_reconstruction}
\end{table}

\begin{figure}[h]
    \centering
    \includegraphics[width=0.5\linewidth]{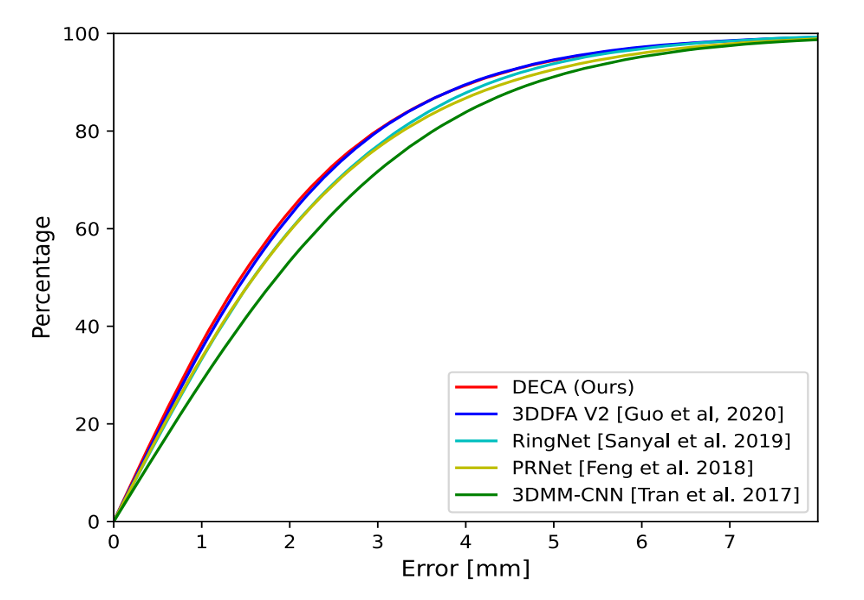}
    \caption{LQ curve}
    \label{fig:enter-label}
\end{figure}
\begin{figure}[h]
    \centering
    \includegraphics[width=0.5\linewidth]{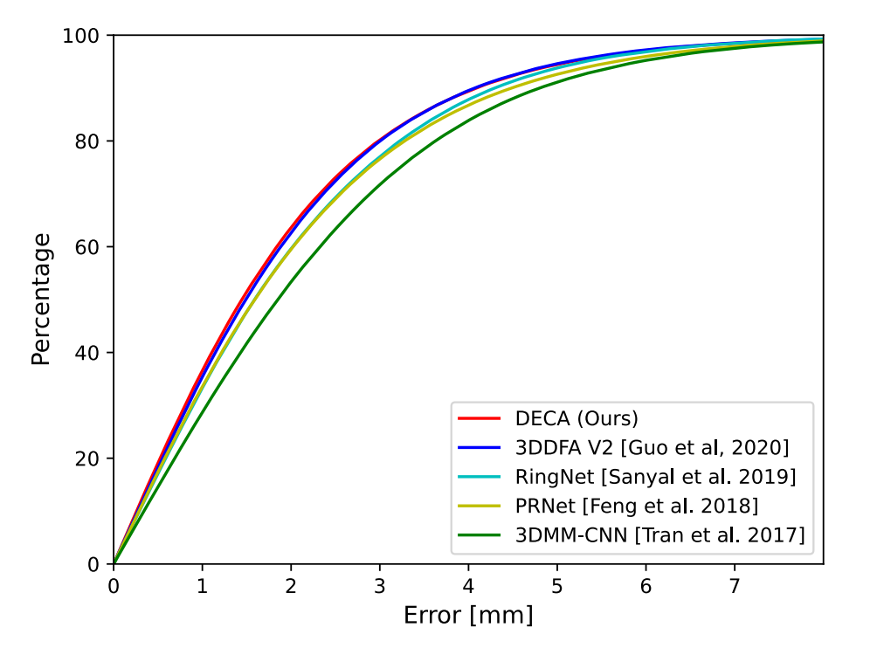}
    \caption{HQ curve}
    \label{fig:enter-label}
\end{figure}

\begin{figure}[H]
    \centering
    \includegraphics[width=1\linewidth]{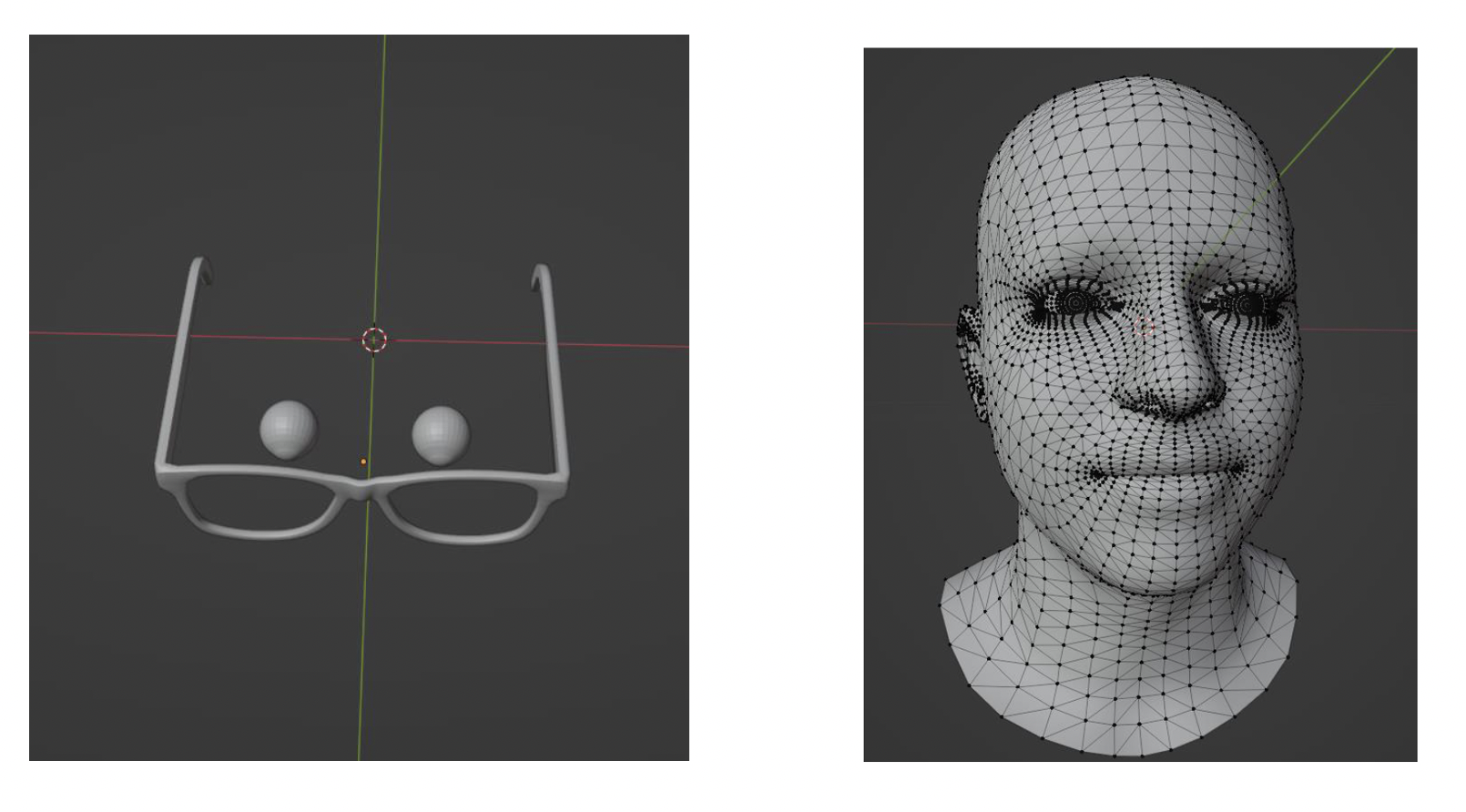}
    \caption{3D meshes of glass and head}
    \label{fig:enter-label}
\end{figure}

\begin{figure}[H]
    \centering
    \includegraphics[width=1\linewidth]{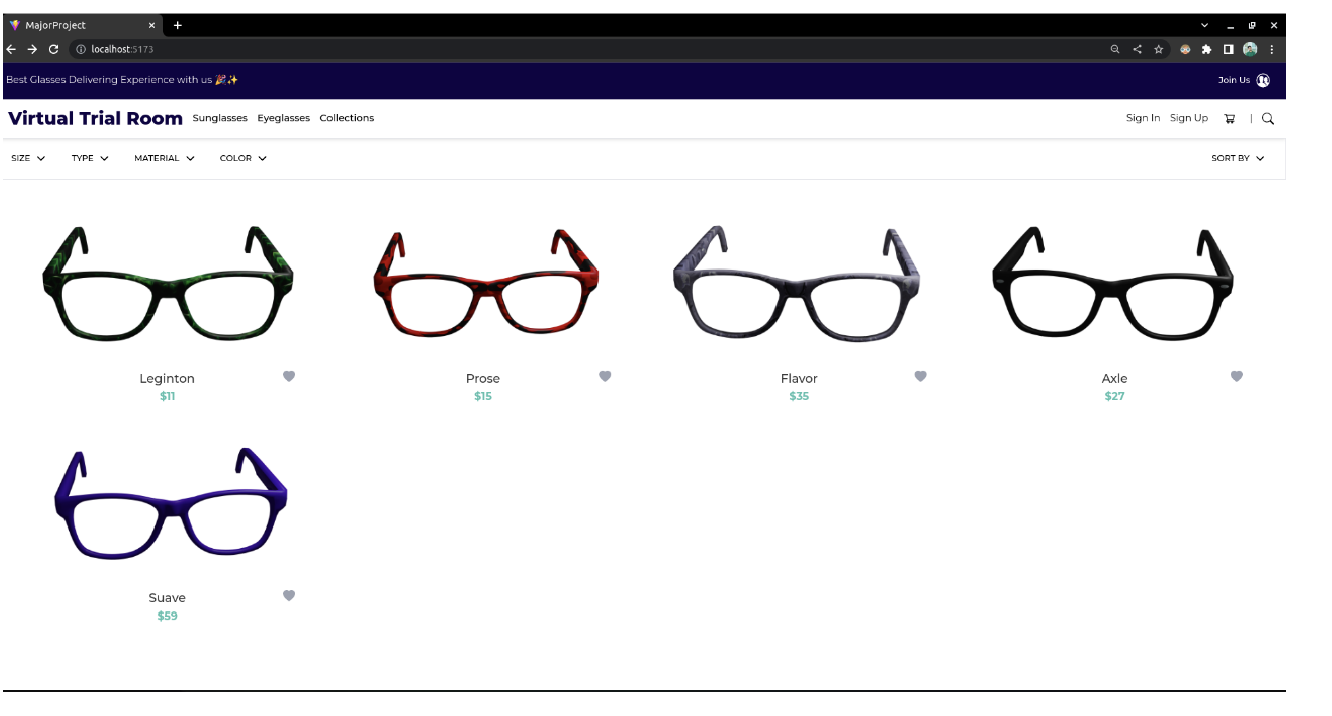}
    \caption{Virtual store}
    \label{fig:enter-label}
\end{figure}
\section{CONCLUSION}
The primary objective of the project "Virtual Trial Room with Computer Vision and Machine Learning" was to tackle the real-life issue pertaining to eyeglasses. The project aimed to convert a 2D image of a person's head into a 3D model, create multiple 3D model of eyeglasses, and integrate them together. After dedicated efforts, the project has made significant strides towards achieving its goals. The team has successfully generated a 3D head model of a person from a single 2D image and created a 3D model of eyeglasses utilizing Blender. Moreover, both 3D models have been integrated and showcased to users via a user-friendly website. The website permits the user to upload a single image, and the 3D model of the head with a fitted eyeglass is automatically generated. The website offers multiple 3D eyeglass models, giving users the flexibility to try on different styles. In conclusion, the developed system enables users to easily check the compatibility of eyeglasses with their facial structure by uploading a single image, eliminating the need for a physical test.
In conclusion, project “Virtual Trial Room with Computer Vision and Machine Learning was able to achieve its objective”. 

\section*{Acknowledgment}
We extend our heartfelt gratitude to our project supervisor, Er. Umesh Kanta Ghimire, for his invaluable guidance throughout this project. We also appreciate the support and suggestions from the Department of Electronics and Computer Engineering at Thapathali Campus, which were instrumental in the selection and execution of this project. Lastly, we would like to thank everyone who contributed, directly or indirectly, to the successful completion of this research.

\authorbio{tulashi}{Tulashi Prasad joshi }{Tulashi Prasad Joshi is a Research
and Development Engineer
working mainly on Computer Vision, Deep Learning and 3D. He holds a Bachelors
Degree in Electronics, Communication, and Information
Engineering from Institute
of Engineering, Thapathali Campus,
Tribhuvan University, Nepal. His research interests focuses
on Deep Learning, Computer Vision, 3D and other
artificial intelligence applications.}\\
\authorbio{amrendra}{Amrendra Kumar Yadav }{Amrendra Kumar Yadav is a Software Engineer specializing in Computer Vision, Artificial Intelligence, Cloud Computing, and Software developoment. He graduated with a Bachelor's Degree in Electronics, Communication, and Information Engineering from the Institute of Engineering, Thapathali Campus, Tribhuvan University, Nepal. his research interests center around the application of AI, cloud, and computer vision.}\\
\authorbio{arjun}{Arjun Chhetri }{Arjun Chhetri an Electronics Communication and Information Engineer graduated from Institute of Engineering Thapathali Campus, Tribhuwan University. His research interest includes, web technologies, cycber security and cloud computing.}\\
\authorbio{suraj}{Suraj Agrahari }{Suraj Agrahari an Electronics Communication and Information Engineer graduated from Institute of Engineering Thapathali Campus, Tribhuwan University. His research interest includes, web technologies, computer vision, Data science.}\\
\authorbio{ukg}{Umesh Kanta Ghimire }{Umesh Kanta Ghimire is Head of Department at Department of Electronics and Computer Engineering, IOE, Thapathali Campus, Tribhuwan University. His research interest includes, machine learning, image processing, natural language processing and IoT}

\begin{thebibliography}{99} 

\bibitem{ref1}
Y. Feng, H. Feng, M. J. Black, and T. Bolkart. 
"Learning an Animatable Detailed 3D Face Model from In-The-Wild Images." 
Max Planck Institute for Intelligent Systems and Max Planck ETH Center for Learning Systems, Germany, 2018.

\bibitem{ref2} 
Shreya Kamani, Neel Vasa, Kriti Srivastava, "Virtual trial room using augmented reality", \textit{International Journal of Advanced Computer Technology (IJACT)}, Vol. 3/6, Dec. 2014, pp. 98-102.

\bibitem{ref3} 
F. Isikdogan and G. Kara, “A Real Time Virtual Dressing Room Application using Kinect”, Cmpe537 Computer Vision Course Project, Bogazici University, January 2012.

\bibitem{ref4} 
Cecilia Garcia Martin, Erdal Oruklu, "Human Friendly Interface Design for Virtual Fitting Room Applications on Android Based Mobile Devices", \textit{Journal of Signal and Information Processing}, Vol. 3/4, 2012, pp. 481-490.

\bibitem{ref5} 
Nikki Singh, Sagar Murade, Prem Lone, Vikas Mulaje "Virtual Trial Room", \textit{Vishwakarma Journal of Engineering Research}, Volume 1 Issue 4, December 2017.

\bibitem{ref6} 
G. van Rossum and F. L. Drake. 
\textit{Python 3 Reference Manual}. 
Python Software Foundation, 2009. Available: \url{https://www.python.org}

\bibitem{ref7} 
Saurabh Botre, Sushant Chaudhari, Shamla Mantri, "Virtual Trial Room", \textit{International Journal of Computer Science Trends and Technology (IJCST)}, Volume 2 Issue 2, Mar-Apr 2014.

\bibitem{ref8} 
Y. Feng, F. Wu, X. Shao, Y. Wang, and X. Zhou. 
"Joint 3D Face Reconstruction and Dense Alignment with Position Map Regression Network." 
In \textit{Proceedings of the European Conference on Computer Vision (ECCV)}, 2018.


\bibitem{ref9} 
Vipin Paul, Sanju Abel J., Sudharsan S., Praveen M, "VIRTUAL TRIAL ROOM", \textit{South Asian Journal of Engineering and Technology}, Vol.3, No.5 (2017), pp. 87–96.

\bibitem{ref10} 
Jamal Dargham, Ali Chekima, “Lips Detection in the Normalised RGB Colour Scheme”, \textit{Information and Communication Technologies, ICTTA '06. 2nd}, Volume: 1, 2006.

\bibitem{ref11} 
Y. Lin, Mao-Jiun and J. Wang, “Automated body feature extraction from 2D images”, \textit{Expert Systems with Applications}, vol. 38, no. 3, pp. 2585-2591, 2011.

\bibitem{ref12} 
P. J. Phillips, H. Moon, S. A. Rizvi, and P. J. Rauss, “The FERET evaluation methodology for face recognition algorithms,” \textit{IEEE Trans Pattern Anal. Mach. Intell.}, vol. 22, no. 10, pp. 1090–1104, Oct. 2000.

\bibitem{ref13} 
Q. Cao, L. Shen, W. Xie, O. M. Parkhi, and A. Zisserman. 
"VGGFace2: A Dataset for Recognising Faces across Pose and Age." 
In \textit{2018 13th IEEE International Conference on Automatic Face and Gesture Recognition (FG 2018)}, Xi'an, China, May 15-19, 2018, pp. 67-74. 
doi:10.1109/FG.2018.0002.

\bibitem{ref14} 
Y. Zhang and W. Deng. 
"Class-Balanced Training for Deep Face Recognition." 
In \textit{Proceedings of the IEEE/CVF Conference on Computer Vision and Pattern Recognition Workshops (CVPRW)}, 2020, pp. 824--825.

\bibitem{ref15}
A. Nagrani, J. S. Chung, W. Xie, and A. Zisserman. 
"VoxCeleb: Large-scale Speaker Verification in the Wild." 
\textit{Computer Science and Language}, 2019.

\bibitem{ref16}
J. S. Chung, A. Nagrani, and A. Zisserman. 
"VoxCeleb2: Deep Speaker Recognition." 
In \textit{Proceedings of Interspeech}, 2018.

\bibitem{ref17}
A. Nagrani, J. S. Chung, and A. Zisserman. 
"VoxCeleb: A Large-scale Speaker Identification Dataset." 
In \textit{Proceedings of Interspeech}, 2017.

\bibitem{ref18} 
T. Li, T. Bolkart, M. J. Black, H. Li, and J. Romero. Learning
a model of facial shape and expression from 4D scans. ACM
Transactions on Graphics, 36(6):194, 2017

\bibitem{ref19} 
A. Weiss, "Chumpy: A Python Library for Optimization and Automatic Differentiation." 
Available: \url{https://github.com/mattloper/chumpy}

\bibitem{ref20} 
M. Abadi, P. Barham, J. Chen, et al. 
"TensorFlow: A System for Large-Scale Machine Learning." 
In \textit{Proceedings of the 12th USENIX Symposium on Operating Systems Design and Implementation (OSDI)}, Savannah, GA, USA, 2016, pp. 265-283. 
Available: \url{https://www.tensorflow.org}

\bibitem{ref21} 
C. Cao, Y. Weng, S. Zhou, Y. Tong, and K. Zhou. 
"FaceWarehouse: A 3D Facial Expression Database for Visual Computing." 
\textit{IEEE Transactions on Visualization and Computer Graphics}, vol. 20, no. 3, pp. 413-425, 2014.

\bibitem{ref22} 
P. Paysan, R. Knothe, B. Amberg, S. Romdhani, and T. Vetter. 
"A 3D Face Model for Pose and Illumination Invariant Face Recognition." 
In \textit{Proceedings of the IEEE International Conference on Advanced Video and Signal-Based Surveillance (AVSS)}, Genova, Italy, 2009, pp. 296-301.

\bibitem{ref23} 
Blender Online Community. 
"Blender - A 3D Modelling and Rendering Package." 
Blender Foundation, Stichting Blender Foundation, Amsterdam, 2023. Available: \url{https://www.blender.org}

\bibitem{ref24} 
J. Guo, X. Zhu, Y. Yang, F. Yang, Z. Lei, and S. Z. Li. 
"Towards Fast, Accurate and Stable 3D Dense Face Alignment." 
In \textit{Proceedings of the European Conference on Computer Vision (ECCV)}, 2020.

\bibitem{ref25} 
J. Guo, X. Zhu, and Z. Lei. 
"3DDFA." 
Available: \url{https://github.com/cleardusk/3DDFA}, 2018.

\bibitem{ref26} 
S. Sanyal, T. Bolkart, H. Feng, and M. J. Black. 
"Learning to Regress 3D Face Shape and Expression from an Image without 3D Supervision." 
In \textit{Proceedings of the IEEE/CVF Conference on Computer Vision and Pattern Recognition (CVPR)}, Long Beach, CA, USA, 2019, pp. 7763--7772.

\bibitem{ref27} 
A. T. Tran, T. Hassner, I. Masi, and G. Medioni. 
"Regressing Robust and Discriminative 3D Morphable Models with a Very Deep Neural Network." 
In \textit{Proceedings of the IEEE/CVF Conference on Computer Vision and Pattern Recognition (CVPR)}, 2017.

\bibitem{ref28} 
A. T. Tran, T. Hassner, I. Masi, E. Paz, Y. Nirkin, and G. Medioni. 
"Extreme 3D Face Reconstruction: Seeing Through Occlusions." 
In \textit{Proceedings of the IEEE/CVF Conference on Computer Vision and Pattern Recognition (CVPR)}, 2018.

\end{thebibliography}
\end{document}